\documentclass[letterpaper, 10 pt, conference]{ieeeconf}
\usepackage{amsmath,amsfonts}
\usepackage{amssymb}
\usepackage{algorithm}
\usepackage{algpseudocode}
\usepackage{array}
\usepackage[caption=false,font=normalsize,labelfont=sf,textfont=sf]{subfig}
\usepackage{textcomp}
\usepackage{stfloats}
\usepackage{url}
\usepackage{verbatim}
\usepackage{graphicx}
\usepackage{cite}
\usepackage{orcidlink}
\usepackage{xcolor}
\usepackage{bbold}
\usepackage{multirow}
\usepackage{booktabs}
\let\labelindent\relax
\usepackage{enumitem}
\usepackage{pgfplots}
\pgfplotsset{compat=1.18}
\usepackage{pgfplotstable}
\usepgfplotslibrary{groupplots}
\definecolor{baseColor}{HTML}{F3F3F3}
\definecolor{baseBorder}{HTML}{D9D9D9}
\definecolor{feColor}{HTML}{FF7C80}
\definecolor{feBorder}{HTML}{FF4F53}
\definecolor{rtColor}{HTML}{84E291}
\definecolor{rtBorder}{HTML}{29B13C}

\newif\ifshownotes
\shownotesfalse

\PassOptionsToPackage{bookmarks=false}{hyperref}
\hypersetup{hidelinks}

\IEEEoverridecommandlockouts
\begin{document}
\bstctlcite{MyBSTcontrol}

\title{\LARGE \bf
FlowCorrect: Efficient Interactive Correction of Generative Flow Policies for Robotic Manipulation
}

\author{Edgar Welte, Yitian Shi, Rosa Wolf, Maximillian Gilles, Rania Rayyes
\thanks{Institute for Material Handling and Logistics (IFL), Karlsruhe Institute of Technology, 76131, Germany. Email: \texttt{edgar.welte@kit.edu}. \newline This work is supported by the German Federal Ministry of Research, Technology, and Space (BMFTR) under the Robotics Institute Germany (RIG), the DFG SFB-1574-471687386 project, and the Ministry of Science, Research and Arts of the Federal State of Baden-Württemberg within the InnovationCampus Future Mobility.}
}

\maketitle

\vspace{1em}

\begin{abstract}
Generative manipulation policies can fail catastrophically under deployment-time distribution shift, yet many failures are near-misses: the robot reaches almost-correct poses and would succeed with a small corrective motion. We propose FlowCorrect, a modular interactive imitation learning approach that enables deployment-time adaptation of flow-matching manipulation policies from sparse, relative human corrections without retraining. During execution, a human provides brief corrective pose nudges via a lightweight VR interface. FlowCorrect uses these sparse corrections to locally adapt the policy, improving actions without retraining the backbone while preserving the model performance on previously learned scenarios. We evaluate on a real-world robot across four tabletop tasks: pick-and-place, pouring, cup uprighting, and insertion. With a low correction budget, FlowCorrect achieves an 80\% success rate on previously failed cases while preserving performance on previously solved scenarios. The results clearly demonstrate that FlowCorrect learns from very few demonstrations and enables fast, sample-efficient, incremental, human-in-the-loop corrections of generative visuomotor policies at deployment time in real-world robotics.

\end{abstract}

\section{INTRODUCTION}
Recent years have witnessed major progress in large-scale imitation learning. These advances have produced foundational behavior models, including Vision-Language-Action (VLA) models~\cite{zhang2024vision, kim2024openvla}, that map rich semantic embeddings to continuous motor commands. In parallel, generative policy learning with diffusion and flow models~\cite{wolf2025diffusion, black2024pi_0, yan2025maniflow} has emerged as a powerful paradigm for action generation. These policies can acquire broad, multimodal manipulation skills from diverse demonstrations. However, real-world deployment remains brittle: out-of-distribution (OOD) situations can cause execution failures, including mistakes or unsafe interactions because test-time states differ from those seen during training~\cite{ross2011Reduction}.

Closing this gap requires not only stronger pre-training—which is often data- and compute-intensive yet still inefficient at covering long-tail OOD pockets—but also mechanisms for continuous, incremental adaptation during deployment~\cite{celemin2022InteractiveImitationLearning}.
A common remedy is parameter-efficient fine-tuning, which adapts pre-trained robot policies to new tasks and embodiments with modest compute~\cite{hu2022lora, kim2025openvlaoft}. Nevertheless, even lightweight fine-tuning typically assumes a relatively stable target distribution and sufficiently representative correction data, which can require extensive expert supervision. In practice, many failures occur in narrow OOD “pockets” of state space and are \emph{near-misses}: the policy reaches an almost-correct state, and a small spatial or temporal adjustment would recover the task. Batch updating on a handful of corrected rollouts can be expensive and brittle, and may induce parameter interference that degrades previously competent behaviors~\cite{zheng2025iManip}. We therefore argue that continual policy learning should support efficient online correction, enabling incremental adaptation to new situations while preserving the base policy's stability and performance.

\begin{figure}
    \centering
    \includegraphics[width=1.0\linewidth]{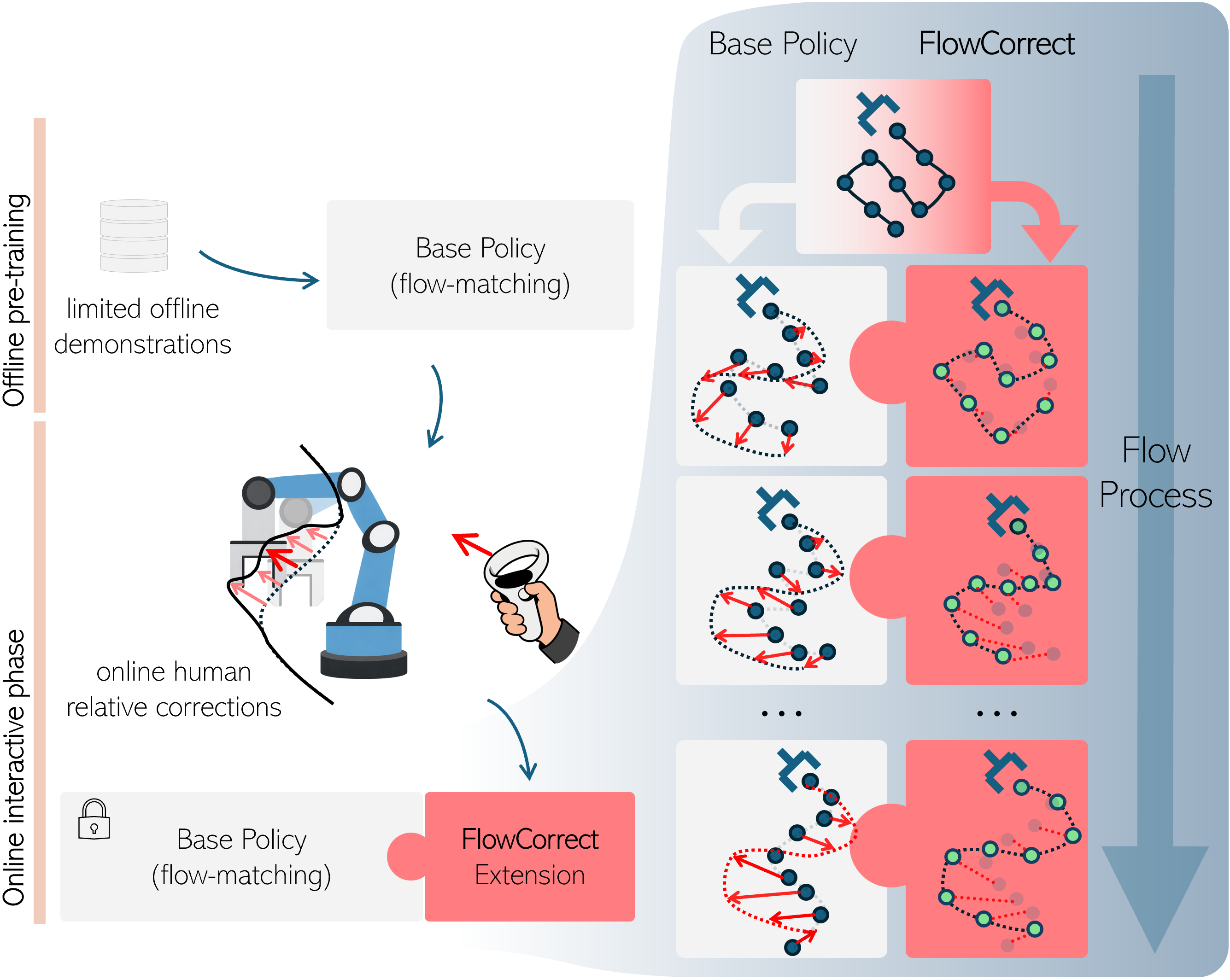}
    \caption{Overview of our interactive \emph{FlowCorrect} framework: A flow matching visuomotor policy is trained from offline demonstrations. During deployment, the frozen base policy runs the robot while a human provides occasional relative corrections. These sparse corrections are used to train our proposed lightweight \emph{FlowCorrect} module that locally steers the policy’s flow field, yielding an adapted policy without retraining the original backbone.}
    \label{fig:teaser}
\end{figure}

Interactive imitation learning (IIL) provides a natural pathway by enabling brief supervisory interventions during execution \cite{welte2025interactive}. This is especially well suited to near-miss failures, where small corrections can convert failure into success without collecting new expert demonstrations. Prior interactive approaches often learn from evaluative feedback (e.g., success/failure or scalar preference~\cite{macglashan2017interactive, christiano2017deep}), which offers limited directional information for precise motion correction. Other approaches~\cite{kellyHGDAggerInteractiveImitation2019, celemin2019interactive} rely on absolute corrections that specify an exact target action, pose, or short trajectory segment. While straightforward, they typically require precise inputs from the human supervisor, increasing cognitive load and often requiring domain expertise.

In this work, we propose \emph{FlowCorrect}, a deployment-time correction framework that enables intuitive, incremental adaptation from sparse human interventions while preserving the broad capabilities of a pre-trained base policy.
We use relative corrections—incremental changes to the robot’s current behavior—rather than requiring full demonstrations or exact target actions. Such corrections are often more natural and intuitive for non-experts~\cite{celemin2015COACH}. We treat these interventions as short-horizon, bounded edits that recover near-miss failures without relearning the underlying skill.

Fig.~\ref{fig:teaser} illustrates our plug-and-play \emph{FlowCorrect} module for flow-based policies: a lightweight correction component that incorporates sparse online human feedback to improve behavior locally, together with a locality-preserving design that leaves the original policy unchanged outside corrected regions. Flow parameterizations are particularly amenable to such edits because they represent behavior as a continuous action flow, enabling targeted local perturbations around encountered states while preserving the global structure learned from offline demonstrations.

To summarize, our contributions are as follows:

\begin{itemize}[leftmargin=*]
    \item \textbf{\emph{FlowCorrect} deployment-time correction for generative manipulation policies}: We introduce an interactive framework for adapting flow-based manipulation policies from sparse human interventions, targeting near-miss failures without full policy retraining.
    \item \textbf{Intuitive human feedback with localized adaptation}: \emph{FlowCorrect} learns from brief relative corrections, localizing updates to corrected situations and preserving the base policy in previously competent regions.
    \item \textbf{Real-robot validation under low correction budgets}: Across four tabletop manipulation tasks, we show that a small number of corrections enables rapid deployment-time recovery of failed cases, improving success while maintaining performance on previously solved scenarios and outperforming full-policy retraining in efficiency.
\end{itemize}

\section{RELATED WORK}
\label{sec:related_work}

\subsection{IIL with human-in-the-loop corrections}
A practical line of IIL improves policies during deployment by allowing a human to intervene when failures are imminent. Pan et al.\cite{pan2025online} introduce Decaying Relative Correction (DRC), where a human supplies a transient relative offset that decays over time, reducing sustained teleoperation burden. The policy is then updated by retraining a diffusion model on corrected demonstrations. Kasaei et al.~\cite{kasaei2023vital_demo} similarly employ relative wrist-delta corrections, but apply them through offline replay with the image-based teleoperation system. They fine-tune the entire policy on a mixture of original and corrected data. In contrast, \emph{FlowCorrect} performs online, incremental updates to a small adapter without accessing the original demonstration set, yielding a more sample-efficient and deployment-friendly adaptation procedure.

In parallel, several systems use absolute interventions where the human takes over control to directly provide actions or trajectory segments, and the policy is updated from the collected intervention data \cite{xu2025hacts,lee2025diffdagger, liuRobotLearningJob2024, chisariCorrectMeIf2022}. For instance, following the recent trend in uncertainty estimation for robotics \cite{shi2025viso, shi2025vmf}, Diff-DAgger \cite{lee2025diffdagger} leverages generative diffusion models to approximate the expert's action distribution and quantify policy divergence, triggering interventions when the robot's predicted trajectory drifts from the learned manifold. However, absolute corrections increase human workload and effectively turn interaction into full teleoperation. \emph{FlowCorrect} instead relies on brief relative pose corrections on a subset of timesteps, keeping human effort low and preserving the distinction between sparse corrective signals and full demonstrations.

\subsection{Residual and modular adaptation from correction data}
To mitigate catastrophic forgetting and focus on learning from failure patterns \cite{shi2024uncertainty}, a common direction is to learn residual or gated modules on top of an existing policy.
Transic \cite{jiang2024transic} learns a residual policy from on-robot human interventions (e.g., using a SpaceMouse) and combines it with a strong base policy to address sim-to-real gaps without overhauling the original controller.
Xu et al.\ \cite{xu2025crdagger} propose Compliant Residual DAgger (CR-DAgger), which collects delta-motion corrections via a compliant end-effector interface and integrates force feedback as an additional modality for contact-rich adaptation.
Following this general idea, \emph{FlowCorrect} also preserves the pre-trained policy and adds a lightweight correction mechanism that activates only when needed, thereby reducing interference outside the narrow OOD regions responsible for failures.
However, unlike these residual controllers, \emph{FlowCorrect} operates on high-level pose predictions from sparse relative corrections and is designed to work with standard hardware and observations.

\subsection{Generative visuomotor policies}
Recent visuomotor policies increasingly utilize generative models to capture multimodal predictions \cite{wolf2025diffusion, shi2025hograspflow, black2024pi_0}, enabling the policy to represent multiple distinct, equally valid action sequences for the same task. Among them, diffusion-based policies \cite{chi2025diffusion} have demonstrated strong performance and flexibility, and are commonly adapted via fine-tuning on additional (corrected) rollouts \cite{pan2025online,lee2025diffdagger}.
Beyond that, flow matching with consistency training \cite{prasad2024consistency} has emerged as a highly data-efficient alternative for high-dimensional control with fast inference. For instance, ManiFlow~\cite{yan2025maniflow} combines flow matching and consistency training to produce dexterous actions in up to 10 steps and shows strong robustness and scaling behavior across diverse manipulation settings.
We build on this generative-policy trend by introducing a correction-oriented module that steers flow in the desired direction while keeping the base flow matching policy.

\section{PROBLEM FORMULATION}
\label{sec:problem-formulation}

\subsection{Imitation Learning Setup}
A robot manipulator with a parallel-jaw gripper operates in discrete control time steps $t\in \mathcal{T} =\{1,...,T\}$. At each step, the robot observes $\boldsymbol{o}_t  = (\boldsymbol{o}_t^{pcd}, \boldsymbol{o}_t^{prio})$: a point cloud $\boldsymbol{o}_t^{pcd} \in \mathbb{R}^{N \times 3}$ of the workspace and its current proprioceptive state $\boldsymbol{o}_t^{prio} \in \mathrm{SE}(3)\times\{0,1\}$; and executes an end-effector pose command $\boldsymbol{a}_t = (\boldsymbol{T}_t, g_t)$, where $ \boldsymbol{T}_t  \in \mathrm{SE}(3)$ denotes the end-effector transformation relative to the robot base and $g_t\in \{0, 1\}$ represents the binary gripper state (open/close), tracked by a low-level Cartesian controller.

A high-level policy $\pi$ maps an observation history to an action sequence\footnote{We denote "$\hat{\boldsymbol{a}}$" as all the \textit{predicted} actions by policies in this work.}:
\[\hat{\boldsymbol{a}}_{t:(t+H-1)}=\pi(\boldsymbol{o}_{(t-K+1):t}),\] where $H$ denotes the prediction horizon and $K$ the length of the observation history.

We assume access to an offline dataset of human teleoperation demonstrations $\mathcal{D}_{\text{demo}}=\{\tau_m\}_{m=1,...,M}$ containing $M$ episodes: $\tau_m=(\boldsymbol{o}_{1:T},\boldsymbol{a}_{1:T})_m$, which we use to train a base policy $\pi_\theta$ parametrized by $\theta$.

\subsection{Preliminaries in Flow Matching Policy}
\label{subsec:fm policy}
In line with ManiFlow~\cite{yan2025maniflow}, we use consistency flow matching (CFM)~\cite{yang2024consistency} as the base policy~$\pi_\theta$, which comprises a visual encoder~$z_\theta$ and a vector field model~$f_\theta$. Specifically, $f_\theta$ parameterizes a latent ordinary differential equation (ODE) in the normalized action space.

Let
$\boldsymbol{x}_n=\{(\boldsymbol{T}^n_h,s^n_h)\}_{h=t}^{t+H-1}
$ denote the subtrajectory at ODE step $n \in \{0, ..., N-1\}$.
During inference, we initialize with Gaussian noise $\boldsymbol{x}_0$ and then iteratively sample:
\[
\boldsymbol{x}_{n+1} = \boldsymbol{x}_n + \Delta k\, f_\theta(\boldsymbol{x}_n, k_n, \boldsymbol{c}),
\]
where $\Delta k = 1/N$ is the step size and $k_n = \Delta k \cdot n$ is the normalized time step. $\boldsymbol{c}=z_\theta(\mathbf{o}_{(t-K+1):t})$ denotes the latent conditioning from the observed point cloud, encoded by $z_{\theta}$. After $N$ inference steps, we interpret:
\[
\hat{\boldsymbol{a}}_{t:(t+H-1)} = \boldsymbol{x}_N
\]
as the obtained action chunks $\hat{\boldsymbol{a}}_{t:(t+H-1)}$.

During training we sample a subtrajectory from a rollout~$\boldsymbol{x}_{\text{gt}} =\boldsymbol{a}_{t:(t+H-1)} \in \tau_m$, noise $\boldsymbol{x}_0$, and continuous flow time step~$k \sim \mathcal{U}(0,1)$. A sample~$\boldsymbol{x} = (1 - k)\boldsymbol{x}_0 + k\boldsymbol{x}_{\text{gt}}$ is defined as the linear interpolation between~$\boldsymbol{x}_0$ and~$\boldsymbol{x}_{\text{gt}}$. The model~$f_\theta$ is trained to predict the velocity~$\boldsymbol{v} = \boldsymbol{x}_{\text{gt}} - \boldsymbol{x}_0$ by minimizing the loss:
\begin{equation}
   \mathcal{L}_\text{F}(\theta) = \mathbb{E}_{(\boldsymbol{x}_0 , \boldsymbol{x}_{\text{gt}}, k)} \left[\left\|f_\theta(\boldsymbol{x}, k, \boldsymbol{c}) - \boldsymbol{v}\right\|_2^2\right].
\end{equation}
In consistency flow matching, an additional objective enforces the velocity predictions at two random flow time steps from a discrete range to be consistent, facilitating smoother velocity predictions according to \cite{yang2024consistency}.

\section{METHODOLOGY}
\label{sec:method}

\subsection{System overview}
Our IIL approach aims to learn an augmented policy $\pi_{\theta+\Delta\theta}$ that combines a frozen base policy $\pi_\theta$ with an additional learnable adapter parameterized by $\Delta\theta$. During online execution, a human teacher monitors the robot and can provide corrective feedback. Specifically, given a rollout generated by the current policy $\pi_{\theta+\Delta\theta}$, the teacher may identify a subset of actions as suboptimal and supply corrected actions online through our teleoperation-based correction interface. These corrections are then used to update the adapter parameters $\Delta\theta$.
\subsection{Objectives for Interactive Corrections}
Let $\mathcal{T}_{\text{corr}} \subseteq\mathcal{T}$ be the set of corrected timesteps, i.e., the subset of the entire execution horizon $\mathcal{T}$. For each $t \in \mathcal{T}_{\text{corr}}$, the teacher specifies a corrected pose $\boldsymbol{a}_t^{\text{corr}} \in \mathrm{SE}(3) \times \{0, 1\}$. We define a binary mask to record the existence of this correction on timestep $t$:
\begin{equation}
     m_t = \mathbb{1}_{\mathcal{T}_{\text{corr}}}(t):=
\begin{cases}
1 & \text{if } t \in \mathcal{T}_{\text{corr}},\\
0 & \text{otherwise}.
\end{cases}
\label{eq:mt}
\end{equation}

We denote the baseline (pre-adaptation) actions by $\hat{\boldsymbol{a}}_t^{\text{base}}$, produced by the frozen base policy $\pi_\theta$, and the adapted (post-adaptation) actions by $\hat{\boldsymbol{a}}_t^{\text{adapt}}$ from $\pi_{\theta+\Delta\theta}$ respectively:
\begin{align*}
  \hat{\boldsymbol{a}}^{\text{base}}_{t:(t+H-1)}&=\pi_{\theta}(\boldsymbol{o}_{(t-K+1):t}); \\
\hat{\boldsymbol{a}}^{\text{adapt}}_{t:(t+H-1)}&=\pi_{\theta+\Delta\theta}(\boldsymbol{o}_{(t-K+1):t}).
\end{align*}

The goal of interactive adaptation is to update the augmented policy $\pi_{\theta+\Delta\theta}$ on the corrected timesteps
$\mathcal{T}_{\text{corr}}$, such that it matches the teacher's corrected actions $\boldsymbol{a}_t^{\text{corr}}$ at timesteps $t\in\mathcal{T}_{\text{corr}}$, while remaining close to the frozen base policy $\pi_\theta$ elsewhere. This is achieved by using only a small number of corrections and a parameter-efficient update. Formally, with base policy parameters $\theta$ frozen, we seek adapter parameters $\Delta\theta$ such that:
\[ \hat{\boldsymbol{a}}_t^{\text{adapt}} \approx
\begin{cases}
\boldsymbol{a}_t^{\text{corr}} & \text{if } m_t = 1,\\
\hat{\boldsymbol{a}}_t^{\text{base}} & \text{otherwise}.
\end{cases}
\]

\begin{figure*}[t]
  \centering
  \includegraphics[width=0.9\textwidth]{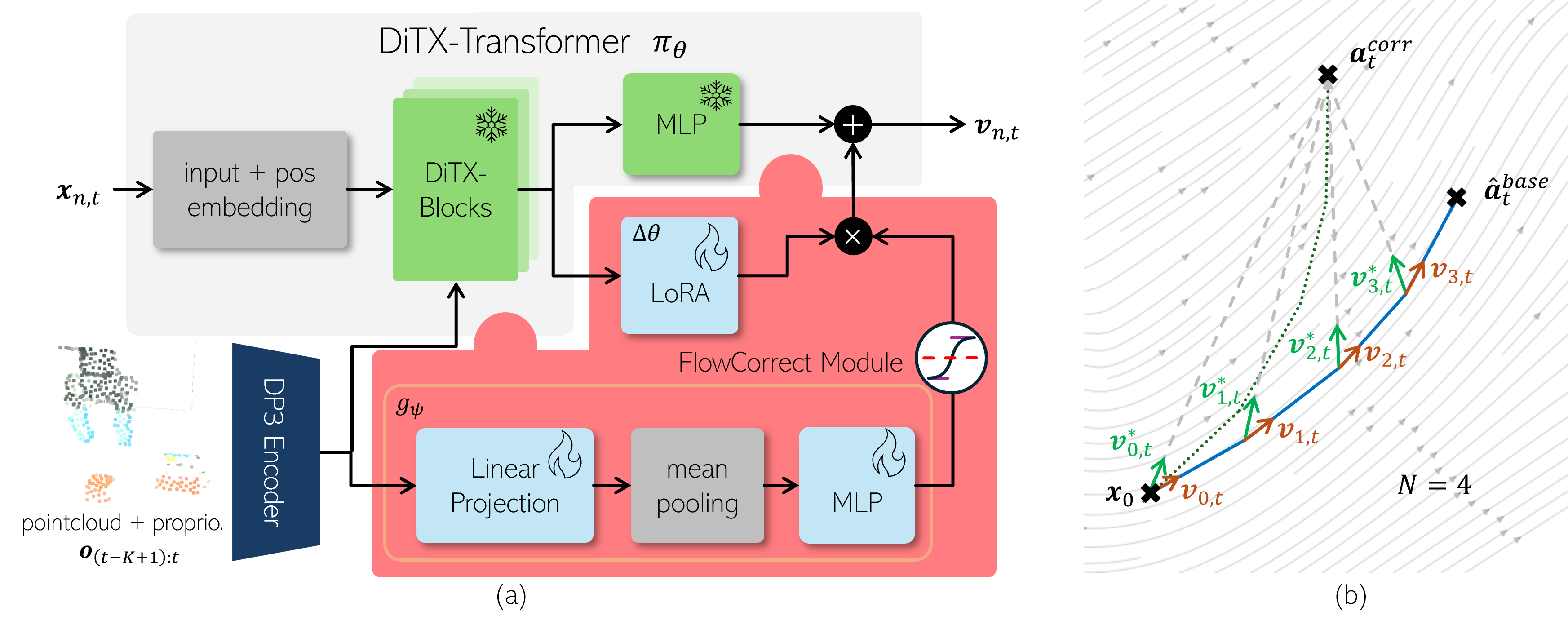}
  \caption{(a) Overview of \emph{FlowCorrect} module that is attached to the DiTX-Transformer from ManiFlow~\cite{yan2025maniflow}: we extend an existing flow matching policy $\pi_{\theta}$ based on DiTX-Transformer with our \emph{FlowCorrect} module. Our lightweight \emph{FlowCorrect} module consists of LoRA adapters (parametrized by $\Delta\theta$) injected into the transformer, and a gating module $g_{\psi}$ that outputs a signal to steer the vector flow field towards the corrected action. (b) Intuition: across N=4 integration steps, \emph{FlowCorrect} iteratively adjusts the predicted velocities from $v_{n,t}$ to $v_{n,t}^*$, steering the rollout from a base action $\hat{\boldsymbol{a}}_t^{\text{base}}$ toward a corrected action $\boldsymbol{a}_t^{\text{corr}}$.}
  \label{fig:arch}
\end{figure*}
\subsection{Interactive Correction Strategy}
\paragraph{Correction Data}
To train the adapter, we collect correction data in a dedicated format during base-policy rollouts. The $i$-th correction sample $\mathcal{S}_i$ consists of: \[\mathcal{S}_i=\{\boldsymbol{o}_{(t-K+1):t }, \hat{\boldsymbol{a}}^{\text{base}}_{t:(t+H-1)},\boldsymbol{a}^{\text{corr}}_{t:(t+H-1)},\boldsymbol{x}_0\}^{(i)},\] each containing observation history $\boldsymbol{o}_{(t-K+1):t}$, base policy actions $\hat{\boldsymbol{a}}^{\text{base}}_{t:(t+H-1)}$, corrected actions $\boldsymbol{a}^{\text{corr}}_{t:(t+H-1)}$, and noise sample $\boldsymbol{x}_0$ used to generate actions.

We additionally record a small set of successful episodes without corrections, which serve as anchor data to discourage global drift of the adapted policy. Since the correction dataset is typically small (i.e., $|\mathcal{T}_{\text{corr}}|\ll|\mathcal{T}| $), we require a highly parameter-efficient adaptation mechanism.

\begin{figure}
    \centering
    \includegraphics[width=1\linewidth]{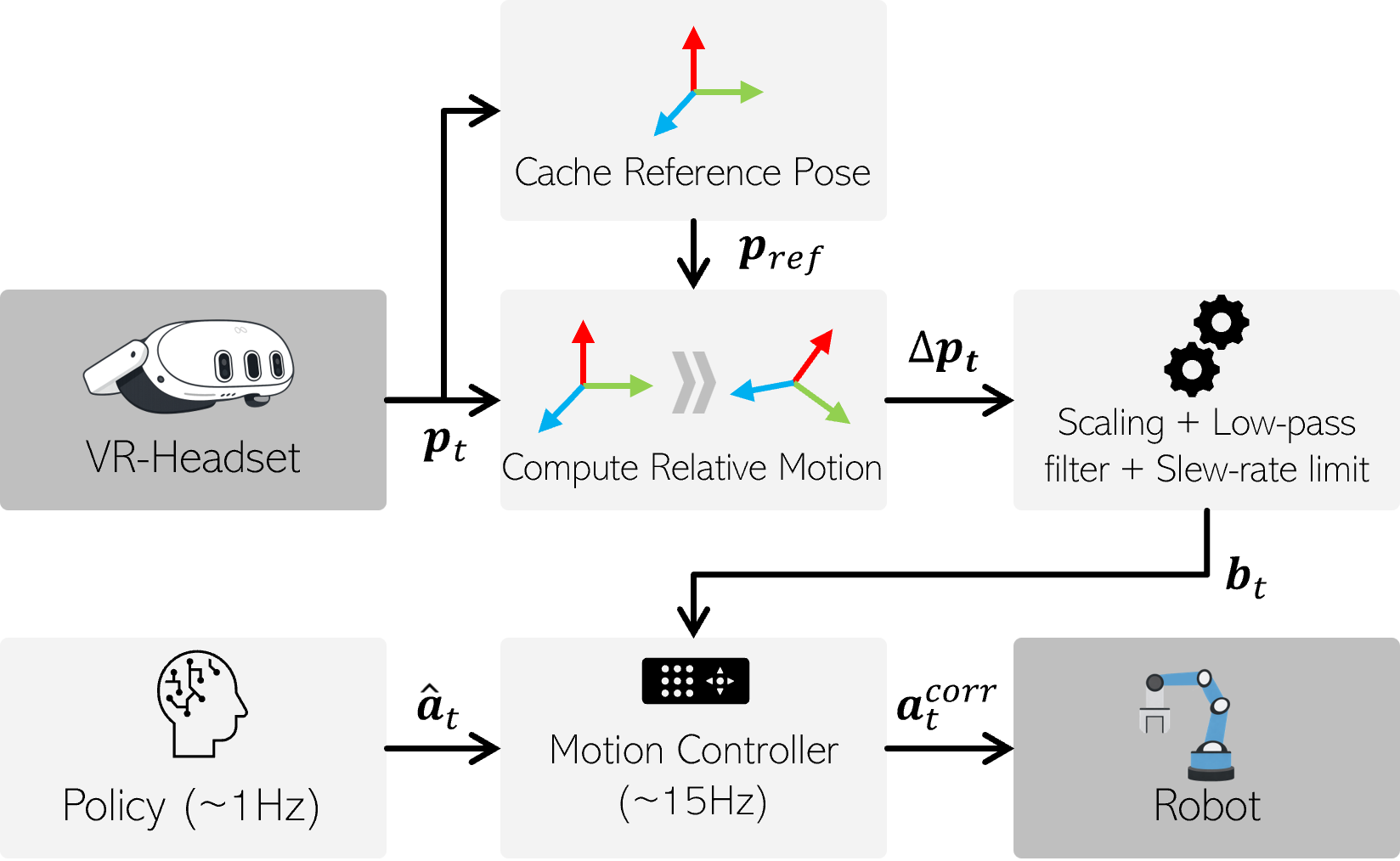}
    \caption{Pipeline of the interactive correction interface.}
    \label{fig:vr_pip}
\end{figure}

We employ relative corrections to enable humans to adjust the robot’s motion during policy execution. In contrast to absolute corrections, in which the human teacher takes full control and teleoperates the robot to specify the complete action, relative corrections consist of a correction offset $\boldsymbol{b}_t $ applied on top of the policy’s nominal output $\hat{\boldsymbol{a}}_t^{base}$:
\[
\boldsymbol{a}_t^\text{corr} = \hat{\boldsymbol{a}}_t^{\text{base}} \oplus \boldsymbol{b}_t
\]
Here, we use $\oplus$ and $\ominus$ to denote group composition and difference in the action space $\mathrm{SE}(3)\times\{0,1\}$ respectively.
\paragraph{Interactive Correction Interface}
We extract a user correction signal through the same VR teleoperation interface used to record demonstrations, as shown in Fig.~\ref{fig:vr_pip}. During policy execution, the user initiates a correction by holding a button on the VR controller. When it is activated, we cache the controller pose $\boldsymbol{p}_{\mathrm{ref}}\in \mathrm{SE}(3)$ as a reference. While the button remains pressed, we compute a 6D correction by decoupling translation and rotation: the translational component is the difference of positions in the world frame, and the rotational component is the relative orientation about world-fixed axes,
\begin{equation*}
\Delta \boldsymbol{p}_t \;=\;
\begin{bmatrix}
\mathbf{t}_t - \mathbf{t}_{\mathrm{ref}}\\
\mathrm{Euler}_{XYZ}\!\left(\mathbf{R}_t\,\mathbf{R}_{\mathrm{ref}}^{\top}\right)
\end{bmatrix},
\end{equation*}
with
{\small
\begin{equation*}
\boldsymbol{p}_t =
\begin{bmatrix}
\mathbf{R}_t & \mathbf{t}_t\\
\mathbf{0}^\top & 1
\end{bmatrix},
\qquad
\boldsymbol{p}_{\mathrm{ref}} =
\begin{bmatrix}
\mathbf{R}_{\mathrm{ref}} & \mathbf{t}_{\mathrm{ref}}\\
\mathbf{0}^\top & 1
\end{bmatrix}.
\end{equation*}
}

This raw correction $\Delta \boldsymbol{p}_t$ is then scaled, low-pass filtered, and slew-rate limited before being applied as an additive offset to the policy action. Attached with the corrective open/close signal $g_t$, we formulate a smoothed target:
\[\tilde{\boldsymbol{b}}_t = \boldsymbol{b}_{t-1} \;\oplus\;\alpha\!\left(\gamma\begin{bmatrix}\Delta \boldsymbol{p}_t \\g_t\end{bmatrix}\ominus \boldsymbol{b}_{t-1}\right),
\quad
\alpha=\frac{dt}{\tau+dt},\]
with scale factor $\gamma$, time constant $\tau$, and control timestep $dt$. To avoid abrupt changes, we additionally limit the maximum step size per timestep via
\[
\boldsymbol{b}_t
=
\boldsymbol{b}_{t-1}
\oplus
\operatorname{clip}\!\left(\tilde{\boldsymbol{b}}_t \ominus \boldsymbol{b}_{t-1},\; r_{\max} dt\right),
\]
where the operator $\operatorname{clip}(\cdot, \rho)$ constrains the magnitude of the relative transformation in $\mathrm{SE}(3)$.
This smooth corrective process yields an intuitive ``nudge'' interface that preserves the overall structure of the policy's behavior while enabling targeted user adjustments. Importantly, the correction loop operates at a higher control rate ($\sim15$Hz) than the policy updates ($\sim1$Hz), thereby improving responsiveness and supporting more natural user interaction.

Practically, we record the correction offset $\boldsymbol{b}_t$ during data collection, corresponding to the action $\boldsymbol{a}_t^\text{corr}$ after it has been fully executed, i.e., when the robot reaches the commanded target pose. Finally, we apply a temporal decay of the correction offset $\boldsymbol{b}_t$ following \cite{pan2025online} after the user releases the button, allowing the robot to smoothly transition back to the policy’s uncorrected output. Our interface produces a full corrected action sequence due to smoothing and decay; thus, $\boldsymbol{a}_t^\text{corr}$ is defined for all time steps.

\subsection{FlowCorrect Module}

To integrate corrections into the augmented policy $\pi_{\theta+\Delta\theta}$, we present \emph{FlowCorrect}, a learnable adapter module based on Low-Rank Adaptation (LoRA) \cite{hu2022lora}. This module is attached to the pre-trained, state-of-the-art ManiFlow~\cite{yan2025maniflow} as the generative policy to enable efficient fine-tuning. Fig.~\ref{fig:arch} (a) illustrates the general architecture of \emph{FlowCorrect}.

Specifically, our core idea is to modify the flow vector field by attaching the LoRA adapter to the MLP head of the DiTX-Transformer in ManiFlow. For a given correction trajectory, we want the flow trajectory starting from the original noise $\boldsymbol{x}_0$ and integrated with the \textit{edited} field to end at the corrected actions $\boldsymbol{a}_{t:(t+H-1)}^{\text{corr}}$ (see Fig.~\ref{fig:arch} (b)).

Let $\boldsymbol{x}_n$ denote the latent action at ODE step $n$ under the edited vector field model $f_{\theta+\Delta\theta}$, the \emph{FlowCorrect} vector field model becomes:
\[
f_{\theta+\Delta\theta}(\boldsymbol{x}_n, k_n, \boldsymbol{c}) = f_\theta(\boldsymbol{x}_n, k_n, \boldsymbol{c}) + \boldsymbol{v}_{\Delta\theta}(\boldsymbol{x}_n, k_n, \boldsymbol{c}),
\]
with observational condition $\boldsymbol{c} = z_{\theta}(\boldsymbol{o}_{(t-K+1):t})$. Restricting LoRA to the head of the DiTX-Transformer and to a low-rank update keeps the number of trainable parameters small ($\approx$10k), and typically yields edits that are localized in hidden-state space.

 For each timestep $t$, we define a per-step target velocity toward the corrected action $\boldsymbol{a}^{\text{corr}}_t$:
\[
\boldsymbol{v}^{*}_{n,t}=\frac{\boldsymbol{a}^{\text{corr}}_t \ominus \boldsymbol{x}_{n,t}}{(N-n)\Delta k},
\]
which can be viewed as the velocity that would exactly reach $\boldsymbol{a}^{\text{corr}}_t$ by the end of the integration if it stayed constant (see Fig.~\ref{fig:arch} (b)). $(N-n)\Delta k$ represents the remaining flow time.

We also introduce a time-dependent weight $w_n$ that emphasizes later steps, as these are more relevant to reach the targeted action:
\[
w_n=\frac{n+1}{N}.
\]
The \emph{FlowCorrect} loss for a single flow trajectory is
\[
\mathcal{L}_{\mathrm{FE}}(\Delta\theta)
=
\frac{1}{N}
\sum_{n=0}^{N-1}
w_n \left\| f_{\theta+\Delta\theta}(\boldsymbol{x}_n, k_n, \boldsymbol{c})_t - \boldsymbol{v}^{*}_{n,t} \right\|_2^2.
\]
In practice, we re-use the logged noise $\boldsymbol{x}_0$ and run the ODE forward with the current $f_{\theta+\Delta\theta}$.
The overall objective over all correction trajectories $\{\mathcal{S}_i\}$ is:
\begin{equation}
\Delta\theta^* = \arg\min_{\Delta\theta}\ \sum_i \mathcal{L}^{(i)}_{\mathrm{FE}}(\Delta\theta).
\label{loss:fe}
\end{equation}
Unlike standard fine-tuning on corrected actions, our objective directly edits the continuous-time flow by matching intermediate ODE velocities to a target field that would reach the corrected trajectory.

Although LoRA is parameter-efficient, its updates can have global effects on the policy. Consequently, improving behavior in one region of the workspace can unintentionally change actions in other regions, potentially reducing performance. To further enforce locality, we introduce a small gating network $g_\psi$ that decides where to apply the flow edit (see Fig.~\ref{fig:arch} (a)):
\[
\alpha_t = g_\psi(\boldsymbol{c}_t) \in [0,1],
\]
where $\boldsymbol{c}_t$ is the observation condition at timestep $t$. The gating network is intentionally small: it first projects $\boldsymbol{c}_t$ into a low-dimensional space via linear projection, aggregates the projected features via mean pooling, and then uses a two-layer Multilayer Perceptron (MLP) to produce the scalar gate $\alpha_t$. The gated \emph{FlowCorrect} vector field model becomes:
\[
f_{\theta+\Delta\theta}(\boldsymbol{x}_n, k_n, \boldsymbol{c}) = f_\theta(\boldsymbol{x}_n, k_n, \boldsymbol{c}) + \alpha_t \boldsymbol{v}_{\Delta\theta}(\boldsymbol{x}_n, k_n, \boldsymbol{c}).
\]

\noindent In general, we train the \textit{FlowCorrect} in two stages:
\begin{enumerate}
  \item Training \emph{FlowCorrect} module: optimize $\Delta\theta$ with Eq.~\ref{loss:fe}, where we fix $\alpha_t \equiv 1$
  \item Gate training: freeze $\theta+\Delta\theta$ and optimize $\psi$ with
    \begin{align*}
\mathcal{L}_G
=& \text{BCE}(\alpha,\, y)
 - \lambda_{\text{ent}}\;\mathrm{H}(\alpha).
\end{align*}

\end{enumerate}
$ \text{BCE}(\alpha,\, y)$ supervises the gate to predict whether an edit should be applied over the current horizon. The ambiguity penalty $\mathrm{H}(\alpha)=\alpha(1-\alpha)$ promotes decisive gating by pushing $\alpha$ toward $0$ or $1$ rather than ambiguous intermediate values. Moreover, we define the ground-truth target $y$ as:
\begin{equation*}
y =
\begin{cases}
1 & \text{if } \exists \, t' \in \{t, \dots, t+H-1\} ,\, t' \in \mathcal{T}_{\text{corr}} \\
0 & \text{otherwise}
\end{cases}
\end{equation*}
which labels the window as positive if any corrected timestep occurs in {\small$t\!:\!t\!+\!H\!-\!1$}. This encourages the gate to open in those cases.

At inference, we apply a threshold $\hat{\alpha}_t=\mathbb{1}[\alpha_t>0.5]$ to obtain a binary ``use edit / do not use edit'' decision per timestep, making the behavior interpretable as applying the learned flow correction only where the human indicated failures during interaction.

\section{EXPERIMENTS}
\label{sec:experiments}
\begin{figure}
    \centering
    \includegraphics[width=1\linewidth]{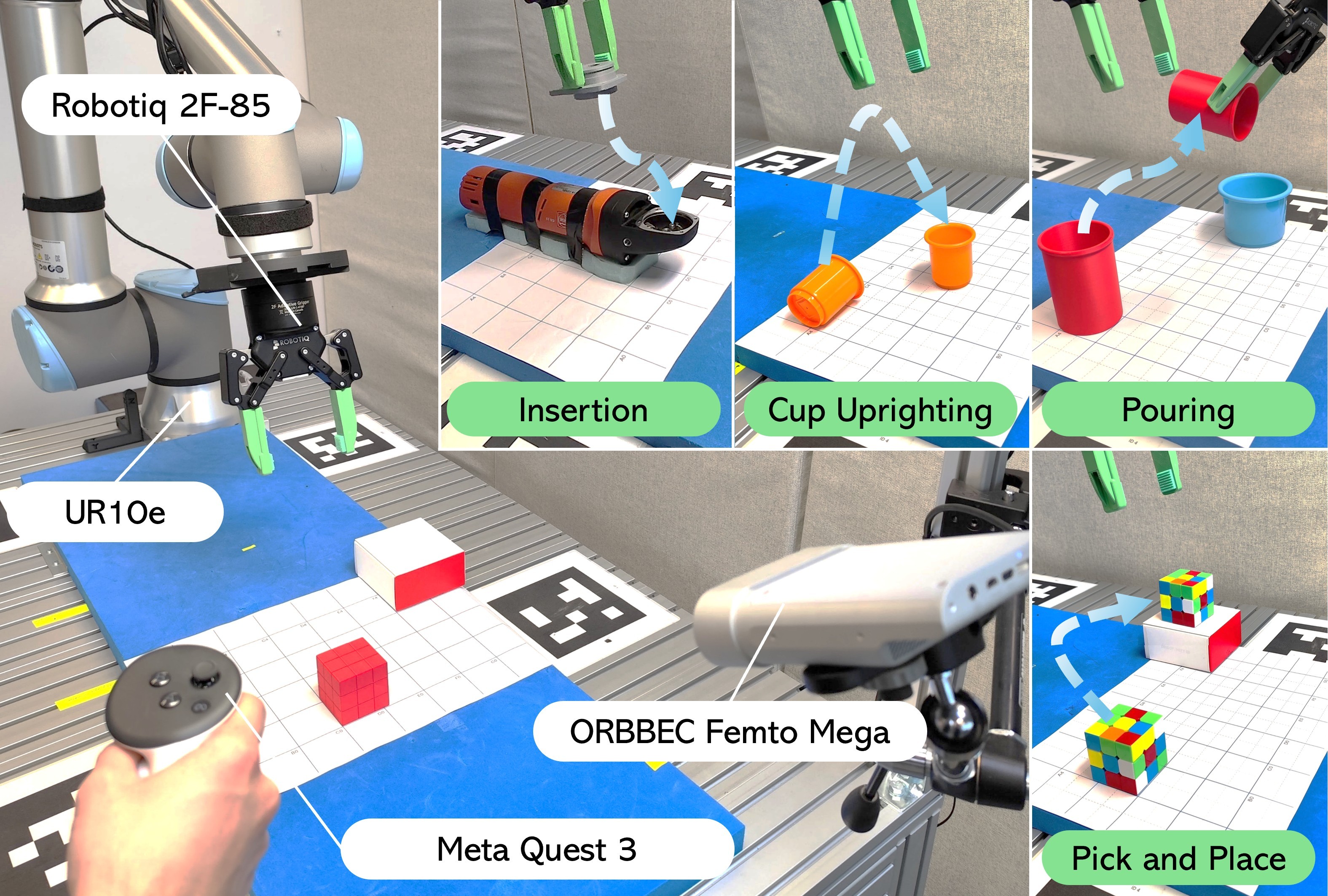}
    \caption{Hardware setup and representative real-world tasks used in the experiments.}
    \label{fig:exp}
\end{figure}

We design real-robot experiments to answer the following questions:

\begin{itemize}
    \item Can \emph{FlowCorrect}, using only local human corrections, reliably fix low-performing situations while preserving the base policy's performance on other states?
    \item How does \emph{FlowCorrect} compare to retraining the entire base policy in terms of performance and efficiency?
    \item What is the impact of our gating mechanism and the usage of uncorrected rollout data?
\end{itemize}

We evaluate our \emph{FlowCorrect} module on four tabletop manipulation tasks illustrated in Fig.~\ref{fig:exp}: (i) Pick-and-Place, (ii) Pouring, (iii) Cup Uprighting, and (iv) Insertion. For each task, we compare three policy types trained with the same backbone and observation/action spaces: (i) the base policy (\textbf{Base}) trained from demonstrations; (ii) our fine-tuned \emph{FlowCorrect} policy (\textbf{FC}) updated from human corrections; (iii) the retrained base policy (\textbf{RT}) updated from the same corrections.

We report success rates over a structured set of in-distribution (ID) and out-of-distribution (OOD) initial conditions, and perform ablations to isolate the contributions of gating and rollout data.

\begin{figure*}[t]
    \centering
    \includegraphics[width=\textwidth]{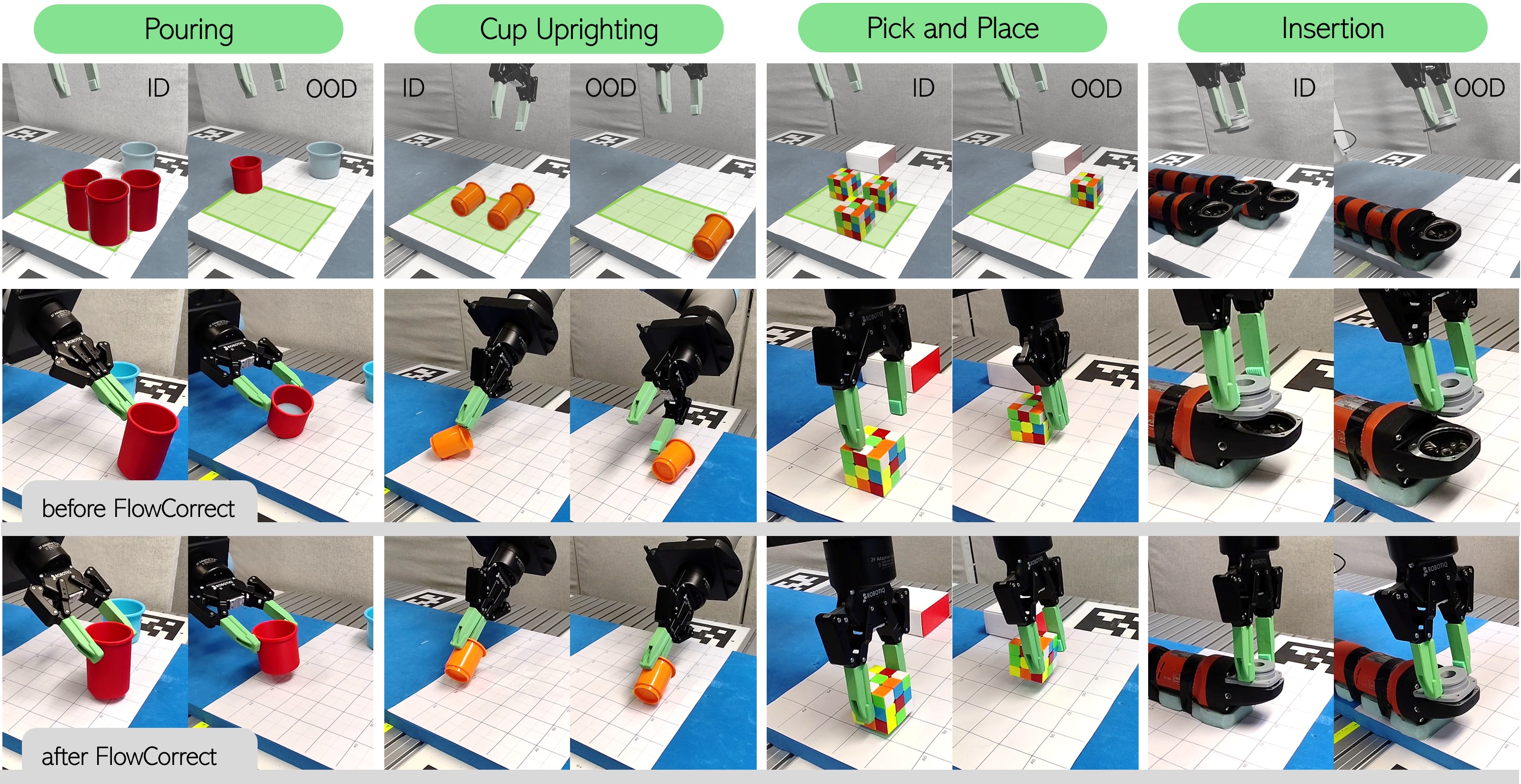}
    \caption{Top row: Selected ID-hard and OOD-hard initial conditions for the four tasks (left to right): Pouring, Cup Uprighting, Pick-and-Place, and Insertion. The green regions indicate the workspace areas covered by the demonstrations. Middle row: Representative failure cases of the base policy under these conditions. Bottom row: Qualitative examples of successful executions after \emph{FlowCorrect} fine-tuning on conditions that previously failed.
    }
    \label{fig:idood}
\end{figure*}

\subsection{Experimental setup}

\paragraph{Hardware and Policy I/O}
All experiments are conducted on a UR10 manipulator equipped with a Robotiq 2F-85 parallel-jaw gripper.
The policy observes $\boldsymbol{o}_t$, including (i) a 3D point cloud $\boldsymbol{o}^{\text{pcd}}_t$ of the workspace captured by an external time-of-flight (ToF) depth camera (Orbbec Femto Mega) and (ii) the robot proprioceptive state $\boldsymbol{o}^{\text{prio}}_t$ (end-effector pose and gripper state).
The policy outputs an absolute 6D end-effector pose command together with a gripper command $\boldsymbol{a}_t$, which are expressed in a common world coordinate frame. Demonstrations and policy rollouts are recorded at 10\,Hz.
We use an observation horizon of $K=2$ timesteps and predict an action sequence of length $H=14$. At execution time, we execute the first $H_{\mathrm{exec}}{=}10$ actions before triggering another inference cycle.

\paragraph{Data collection} For each task, we collect eight expert demonstrations to train the base policy. During deployment, when the base policy fails, a human provides relative corrections for the same failure situation to generate correction rollouts. We also record trajectories from successful base-policy executions. For each selected failure situation, we collect ten corrected rollouts, and we randomly sample five uncorrected rollouts from successful executions.

\paragraph{Tasks} We consider four tasks:
\textbf{Pick-and-Place:} The robot must pick up a Rubik's Cube and place it on a fixed box. The clearance between the cube and the gripper widths is 10~mm.
\textbf{Pouring:} The robot must grasp a cup and perform a pouring motion toward a fixed cup.
\textbf{Uprighting a Cup:} The robot must flip an overturned cup upright and place it in a designated position on the table.
\textbf{Insertion:} The robot must insert a round component into the housing of an angle grinder; the clearance between the component and the housing is 5~mm.
The initial position of all manipulated objects is randomly selected in a workspace area of 15$\times$20~cm.

\paragraph{Policies and Training Variants}
We evaluate three policy types per task: (i) \textbf{Base} policy: trained on $8$ demonstrations. (ii) \textbf{FC} policy (ours): fine-tuned from the base policy by \emph{FlowCorrect} using $10$ corrections for each selected failure case plus $5$ rollout trajectories. (iii) \textbf{\textbf{RT}} policy: retrained (updated) base policy using the same data as \textbf{FC} policy in terms of corrections and rollouts. The \textbf{Base} policy is trained for 3000 epochs, whereas the \textbf{FC} and \textbf{RT} policies are trained for 500 epochs on a single NVIDIA RTX 4090. We use a batch size of 256 for base policy training and 64 for fine-tuning, average runtimes are reported in Table~\ref{tab:resource}.

\subsection{Experiment evaluation}

For each task, we define 30 in-distribution (ID) initial conditions (i.e., object positions) within the workspace region used for demonstration recording. To ensure a more standardized and comparable evaluation, these 30 positions are arranged as a 6×5 grid over the defined workspace. In addition, we define three selected low-performing ID conditions (\emph{ID-hard1/2/3} - identified by evaluation of the base policy), and one selected low-performing OOD condition (\emph{OOD-hard}) to be corrected by a human. In this context, ID conditions, including ID-hard cases, lie within the demonstrated 6×5 workspace grid and therefore represent within-workspace failures rather than OOD extrapolation cases.

The \emph{OOD-hard} condition is selected differently per task. In \emph{Cup Uprighting}, \emph{Pick-and-Place}, and \emph{Insertion}, we chose a random position 3-5 cm outside the workspace as the OOD condition, whereas for the \emph{Pouring} task, we selected a smaller cup height as the OOD condition, since the positional OOD conditions were already robust in that task. Figure~\ref{fig:idood} shows the different ID and OOD conditions as well as some common failure cases of the base policy before adaptation with \emph{FlowCorrect}. Typical failure cases include an unstable grasp, collision with the object, or misalignment with the object. Each ID and OOD condition is evaluated 10 times to define a success rate.

\subsection{Results}

\begin{figure}[t]
\centering
\begin{tikzpicture}
\begin{axis}[
    ybar,
    bar width=8pt,
    width=1.05\columnwidth,
    height=0.55\columnwidth,
    ymin=0, ymax=1.0,
    ymajorgrids,
    grid style={dashed},
    ylabel={Success rate},
    yticklabel style={font=\scriptsize},
    ylabel style={font=\scriptsize},
    symbolic x coords={Insertion, Pick-and-Place, Cup Uprighting, Pouring, Average},
    xtick=data,
    x tick label style={rotate=25, anchor=east, yshift=-3pt, font=\scriptsize},
    enlarge x limits=0.18,
    legend style={
        at={(0.5,1.15)},
        anchor=south,
        legend columns=3,
        draw=none
    },
    nodes near coords,
    point meta=explicit symbolic,
    every node near coord/.append style={
        font=\scriptsize,
        rotate=90,
        anchor=east,
        xshift=1pt
    },
]
\addplot[fill=baseColor, draw=baseBorder] coordinates {
    (Pick-and-Place, 0.5333) []
    (Pouring,       0.6333) []
    (Cup Uprighting,0.5333) []
    (Insertion,        0.3000) []
    (Average,       0.5000) []
};
\addplot[fill=feColor,   draw=feBorder] coordinates {
    (Pick-and-Place, 0.6000) []
    (Pouring,       0.9000) []
    (Cup Uprighting,0.7333) []
    (Insertion,        0.3667) []
    (Average,       0.6500) []
};
\addplot[fill=rtColor,   draw=rtBorder] coordinates {
    (Pick-and-Place, 0.6667) []
    (Pouring,       0.7667) []
    (Cup Uprighting,0.7000) []
    (Insertion,        0.1667) []
    (Average,       0.5750) []
};

\legend{\textbf{Base}, \textbf{FC} (Ours), \textbf{RT}}
\end{axis}
\end{tikzpicture}
\caption{Overall success rate across 30 ID positions in the workspace. Gray denotes the base policy; red denotes our \emph{FlowCorrect} (FC) policy after correcting only three ID conditions; and green denotes the base policy retrained (RT) with the same three ID corrections. Notably, \emph{FlowCorrect} improves the original performance without regression.}
\label{fig:success_bar_id}
\end{figure}
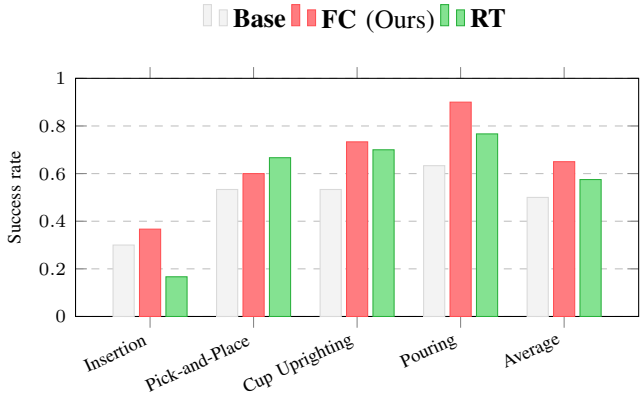

\paragraph{Quantitative results}
Fig.~\ref{fig:success_bar_id} summarizes overall performance across the 30 ID initial conditions.
Across all four tasks, \emph{FlowCorrect} (\textbf{FC}) improves the base policy’s ID success rate, indicating that sparse corrections can generalize beyond the corrected timesteps and stabilize execution in nearby states.
In \textit{Pouring} and \textit{Cup Uprighting}, \textbf{FC} yields large gains. \textit{Pick-and-Place} and \textit{Insertion} require a more precise positioning of the gripper to prevent collisions. Therefore, the improvement does not generalize as strongly here.

Table~\ref{tab:stress_id_ood} reports stress-tests on the selected low-performing ID and OOD conditions.
For \textit{Cup Uprighting}, \textbf{FC} reliably resolves both ID-hard and OOD-hard settings (9--10/10 success across all hard cases).
For \textit{Pouring}, \textbf{FC} fixes all three ID-hard positions (10/10 each), but improves the selected OOD condition only marginally (0/10$\rightarrow$2/10). Notably, this OOD setting corresponds to a \emph{height} change (smaller cup) rather than a positional shift.
For \textit{Pick-and-Place}, \textbf{FC} substantially improves most hard cases (up to 10/10 on ID-hard2 and 9/10 on ID-hard3, and 10/10 on OOD-hard), but one selected ID-hard condition improves only partially (3/10). This ID-hard condition lies spatially close to the chosen OOD condition, and the corresponding corrections are directionally different within a narrow region of the workspace. In such cases, a single locality gate and observation-agnostic LoRA update can lead to \emph{over-correction} toward the OOD solution, effectively overriding the more appropriate edit for the nearby ID case. This highlights an important limitation of local correction schemes when multiple, conflicting edits must coexist at fine spatial granularity.

\begin{table}
\centering
\caption{Stress-test success on selected hard positions. ``ID-hard'' are the three low-performing ID conditions; ``OOD-hard'' is the selected OOD condition.}
\label{tab:stress_id_ood}
\setlength{\tabcolsep}{3pt}
\renewcommand{\arraystretch}{0.95}
\small
\begin{tabular}{llrrrrr}
\toprule
\textbf{Task} & \textbf{Policy} &
\shortstack{\textbf{ID-}\\\textbf{hard1}} &
\shortstack{\textbf{ID-}\\\textbf{hard2}} &
\shortstack{\textbf{ID-}\\\textbf{hard3}} &
\shortstack{\textbf{OOD-}\\\textbf{hard}} &
\textbf{$\Sigma$} \\
\midrule
\multirow{3}{*}{\shortstack[l]{Pick-and-\\Place}}
 & \textbf{Base}      & 0/10 & 0/10 & 0/10 & 0/10 & 0/40\\
 & \textbf{FC} (Ours) & 3/10 & \textbf{10}/10 & 9/10 & \textbf{10}/10 & 32/40\\
 & \textbf{RT}        & \textbf{10}/10 & \textbf{10}/10 & \textbf{10}/10 & \textbf{10}/10 & 40/40\\
\midrule
\multirow{3}{*}{Pouring}
 & \textbf{Base}      & 4/10 & 0/10 & 0/10 & 0/10 & 4/40\\
 & \textbf{FC} (Ours) & \textbf{10}/10 & \textbf{10}/10 & \textbf{10}/10 & 2/10 & 32/40 \\
 & \textbf{RT}        & \textbf{10}/10 & \textbf{10}/10 & \textbf{10}/10 & \textbf{10}/10 & 40/40\\
\midrule
\multirow{3}{*}{\shortstack[l]{Cup\\Uprighting}}
 & \textbf{Base}      & 0/10 & 0/10 & 0/10 & 0/10 & 0/40\\
 & \textbf{FC} (Ours) & \textbf{10}/10 & \textbf{9}/10 & \textbf{10}/10 & \textbf{9}/10 & 38/40\\
 & \textbf{RT}        & \textbf{10}/10 & 8/10 & 8/10 & \textbf{9}/10 & 35/40\\
\midrule
\multirow{3}{*}{Insertion}
 & \textbf{Base}      & 0/10 & 0/10 & 0/10 & 0/10 & 0/40\\
 & \textbf{FC} (Ours) & \textbf{10}/10 & 2/10 & \textbf{10}/10 & \textbf{10}/10 & 32/40\\
 & \textbf{RT}        & 8/10 & \textbf{8}/10 & 7/10 & 9/10 & 32/40\\
\bottomrule
\end{tabular}
\vspace{-1mm}
\end{table}

\paragraph{Qualitative results}
The bottom row of Fig.~\ref{fig:idood} provides representative successful executions after \emph{FlowCorrect} fine-tuning. The examples visually confirm the quantitative improvements shown in Fig.~\ref{fig:success_bar_id} and Table~\ref{tab:stress_id_ood}, demonstrating more stable alignment and execution under challenging initial conditions.

\begin{table}[h]
    \centering
    \caption{Resource usage comparison between our \emph{FlowCorrect} training (\textbf{FC}) and retraining (\textbf{RT}).}
    \begin{tabular}{ccc}
        \hline
        \textbf{Mode} & \textbf{Avg. GPU Memory Usage (GB)} & \textbf{Avg. Runtime (min.)}\\
        \hline
        \textbf{Base} & 18.84$\pm$0.24 & 80.86$\pm$10.01\\
        \textbf{FC} (Ours) & \textbf{4.35$\pm$0.15} & \textbf{30.24$\pm$5.45} \\
        \textbf{RT} & 19.23$\pm$0.25 & 52.93$\pm$10.96 \\
        \hline
    \end{tabular}

    \label{tab:resource}
\end{table}

\paragraph{Comparison to retraining.}
Retraining (\textbf{RT}) achieves consistently strong performance on the hard cases.
However, \textbf{FC} is largely competitive with \textbf{RT} in overall ID success (Fig.~\ref{fig:success_bar_id}) while updating only a small LoRA module and a lightweight gate.
Notably, in \textit{Insertion}, \textbf{RT} shows a substantial drop in overall ID success, suggesting that full-model updates can be sensitive in high-precision settings and may shift behavior in previously successful regions.
In addition, \textbf{RT} incurs a substantially larger training-time footprint (GPU memory and runtime; Table~\ref{tab:resource}), whereas \textbf{FC} provides a more deployment-friendly adaptation mechanism.

\begin{table}[t]
\centering
\caption{Ablation study on \emph{FlowCorrect}. Report success (in \%) on 30 ID positions across all tasks, after applying corrections to four hard cases.}
\label{tab:ablation}
\begin{tabular}{lcccc}
\hline
\textbf{Variant} & \textbf{ID-30 Avg} & \textbf{ID-hard Avg} & \textbf{OOD-hard Avg} \\
\hline
\textbf{Base} & 49.76 & 3.33 & 0.00 \\
\textbf{FC} (full) & \textbf{65.00} & 85.83 & 77.50 \\
\textbf{FC} w/o gate & 54.17 & 69.17 & 75.00 \\
\textbf{FC} w/o rollouts & 55.84 & 80.00 & 87.50 \\
\textbf{RT} & 57.50 & \textbf{90.83} & \textbf{95.00} \\
\textbf{RT} w/o rollouts & 36.67 & \textbf{90.83} & 87.50 \\
\hline
\end{tabular}
\vspace{1mm}
\end{table}

\paragraph{Ablation insights.}
Table~\ref{tab:ablation} shows that the gating mechanism is critical for preserving ID performance: removing the gate drops ID-30 success from 65.00\% to 54.17\%, consistent with the gate preventing unintended global drift.
Using a small set of uncorrected rollouts also improves stability: training \textbf{FC} without rollouts reduces ID-30 performance (55.84\%) and changes the trade-off between hard-case gains and generalization, indicating that anchor trajectories help maintain the base policy’s behavior outside corrected regions. Anchor rollouts regularize toward the base behavior; this can reduce adaptation strength on OOD conditions, explaining the higher OOD-hard score without rollouts for \textbf{FC}.

\subsection{Discussion and Outlook.}
Our experiments reveal two key failure modes. First, when multiple hard cases with conflicting required corrections lie within a tight spatial neighborhood, a single global LoRA edit combined with a coarse gate can induce interference, leading to over-correction toward one regime. Second, OOD shifts driven by object geometry are less effectively addressed by pose-centric local edits, suggesting that the current correction pathway is not selective enough with respect to scene-specific cues.
Due to limited time, we focus our analysis on four representative corrections, each targeting hard ID and OOD cases across the four tasks studied in this paper.
We envision observation-conditioned edits, multiple lightweight experts with learned routing, or a small memory of localized patches to reduce interference between nearby correction regimes. Finally, the current gate supervision labels an entire horizon as positive if any timestep is corrected; future work should explore finer-grained, per-timestep gate targets to limit the temporal extent of edits and reduce unintended drift.

\section{CONCLUSIONS}
\label{sec:conclusion}

We presented \emph{FlowCorrect}, an interactive deployment-time adaptation approach for flow-matching manipulation policies that targets near-miss failures.
\emph{FlowCorrect} enables intuitive incremental human-in-the-loop corrections.
It keeps the pre-trained backbone frozen and learns a lightweight LoRA-based correction module that locally steers the policy’s flow-field from sparse \emph{relative} human nudges.
Central to our approach is a flow-edit objective that steers intermediate ODE velocities toward corrected trajectories, enabling rapid repair without retraining the base policy. Across four real-robot tabletop tasks, \emph{FlowCorrect} substantially improves hard-case success under low correction budgets, while preserving or improving overall in-distribution performance and requiring far less training-time compute than full retraining. Future work will focus on observation-conditioned edits and finer-grained routing to better resolve nearby, conflicting corrections and shifts driven by object geometry.

\bibliographystyle{IEEEtran}
\bibliography{bibliography}

\vfill

\end{document}